\documentclass[letterpaper, 10pt, conference]{ieeeconf}
\IEEEoverridecommandlockouts
\usepackage{enumerate}
\usepackage{graphics} 
\usepackage{epsfig} 
\usepackage{mathptmx} 
\usepackage{times} 
\usepackage{amsmath} 
\usepackage{amssymb}  
\usepackage{subfigure}
\usepackage{array}
\usepackage{color}
\usepackage{cite}
\usepackage{multirow}

\newcommand{\eg}{\textit{e.g.,}~}

\newcommand{\etal}{\textit{et al.}~}
\newcommand{\etc}{\textit{etc.}}

\newcommand{\B}{{\tt BLOCKED}}
\newcommand{\F}{{\tt FREE}}
\newcommand{\ud}{{\tt UNDEFINED}}

\newcommand{\argmax}{\operatornamewithlimits{argmax}}

\usepackage{hyperref}

\title{\LARGE \bf
	End-to-end Learning of Image based Lane-Change Decision
}

\author{Seong-Gyun Jeong, Jiwon Kim, Sujung Kim, and Jaesik Min%
\thanks{The authors are with autonomous driving team at NAVER LABS Corp., Republic of Korea 
\{\tt\small{seonggyun.jeong, g1.kim, sujung.susanna.kim, jaesik.min\}@naverlabs.com}}%
}

\begin{document}

\maketitle
\thispagestyle{empty}
\pagestyle{empty}

\begin{abstract}

	We propose an image based end-to-end learning framework that helps lane-%
change decisions for human drivers and autonomous vehicles. The proposed system, 
Safe Lane-Change Aid Network (SLCAN), trains a deep convolutional neural network 
to classify the status of adjacent lanes from rear view images acquired by cameras 
mounted on both sides of the vehicle. Rather than depending on any explicit object
detection or tracking scheme, SLCAN reads the whole input image and directly decides 
whether initiation of the lane-change at the moment is safe or not. We collected and annotated 77,273 rear side view images to train and test SLCAN. Experimental 
results show that the proposed framework achieves 96.98\% classification accuracy 
although the test images are from unseen roadways. We also visualize the saliency 
map to understand which part of image SLCAN looks at for correct decisions.
\end{abstract}

\section{Introduction}
\label{sec:intro}

	Lane-change is a basic driving maneuver that moves the ego-vehicle
into the adjacent lane heading the same direction. On the roadway, we often
encounter situations to change lanes in order to avoid obstacles, overtake  
other vehicles, or merge into traffic. Before initiating lane-change, a driver 
must be aware of his/her surroundings to avoid crash or any other incident. 
For an inexperienced driver, it is a challenging task to simultaneously perceive
traffic of the ego and adjacent lanes. Without concentration, even a skillful 
driver may cause undesired situations during lane-change.

	For safe lane-change decision aids, automakers have developed blind spot detection
(BSD) systems\footnote{The names of technique may vary with the manufacturers, 
\eg BLind Spot Information System (BLIS), Blind Spot Monitoring (BSM), Blind Spot Warning
(BSW), Side Assist, and \etc}. To observe the rear side space, automakers equip a
vehicle with sensors such as cameras or high frequency radars. The BSD system tracks
rear side traffic of the ego-vehicle and warns the driver if the system detects an
object entering the blind spot zone. 

	To make a safe lane-change, autonomous vehicles as well as human drivers
generally perform the following steps: 1) {\it environmental perception} and
2) {\it maneuver decision making}. For the perception task, researchers attempt
to identify driving-relevant objects on the roadway such as vehicles, lanes, and
road markings. Specifically, significant progress has been made on computer
vision based algorithms for object detection and tracking~\cite{Liu2016,Dai2016,Ren2015,Kalal10}. 
For an in-depth review of computer vision applications for intelligent vehicles, 
we refer the reader to~\cite{Sivaraman2013}. In spite of these advances in computer vision algorithms, the use of them may be inadequate for real time applications such as lane-change problem, because there are too many object classes that need to be detected and tracked in street scenes, \eg cars, fence, and trees.
		
	In general, maneuver decision making acts based on the perception results. To predict potential risks for the traffic, the extracted features via perception
algorithms are fused by various schemes, \eg Bayesian networks~\cite{schubert2010}
and fuzzy-related uncertainty representation~\cite{Pellkofer2002}. 
In~\cite{Schlechtrienmen2014}, the most relevant features have been investigated 
with respect to the lane-change intentions in highway scenario. On the other hand,
general driver models for lane-change maneuver have been presented, \eg Foresighted 
Driver Model (FDM)~\cite{Damerow16} and Minimizing Overall Braking Induced by Lane changes
(MOBIL)~\cite{Kesting99}. Although the above mentioned approaches allow a safe
lane-change, complex procedures are required to interpret the situation with respect to lane-change decision.

	\begin{figure}[t]
		\centering      
		\includegraphics[width=9cm]{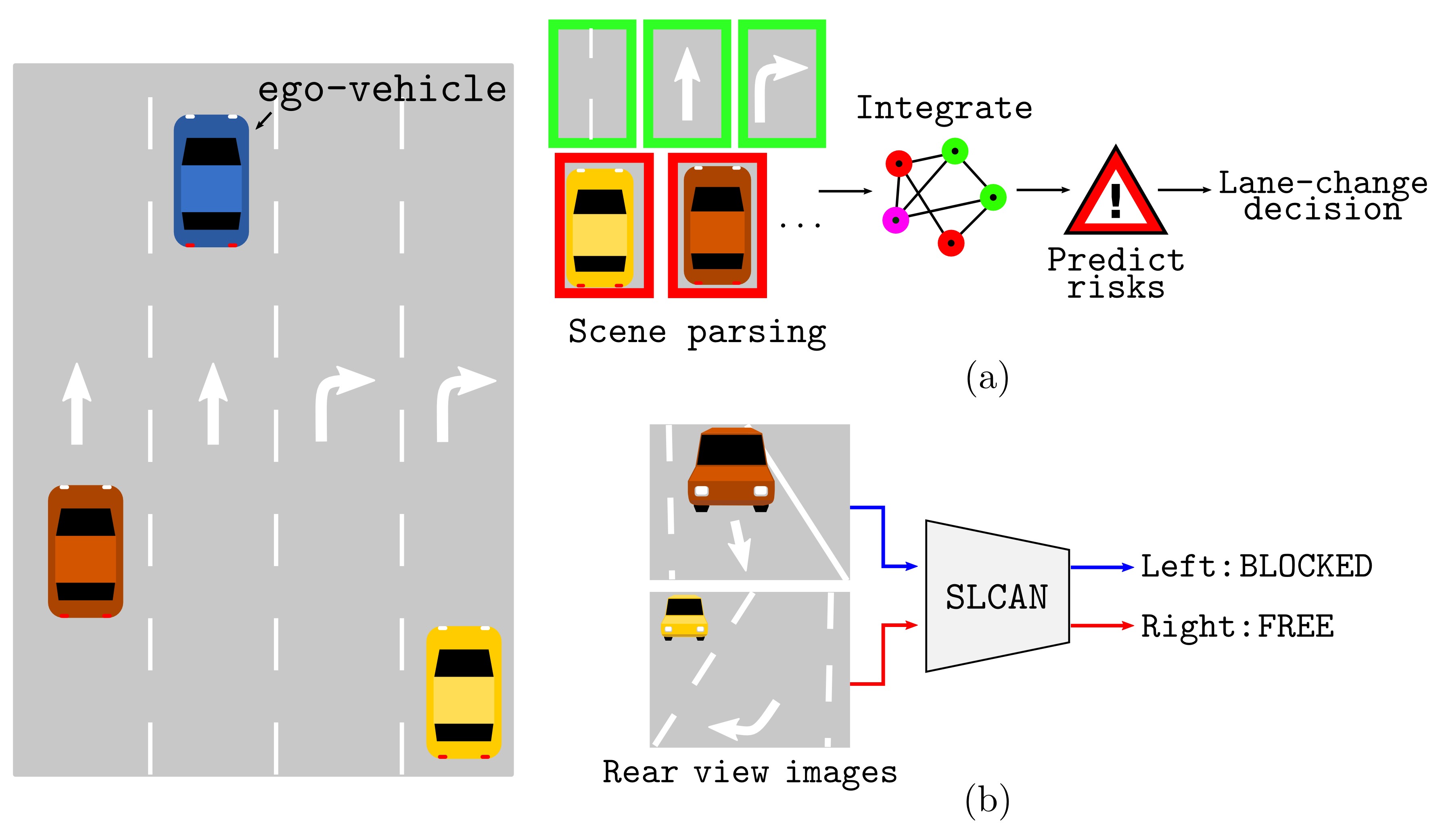}
		\caption{(a) A typical approach to lane-change decision first detects 
individual driving-relevant objects with sophisticated perception algorithms. 
Then, additional steps are needed to integrate the perception results and 
to compute potential risk of lane-change. (b) The proposed approach trains a DCNN 
that interprets rear view images and directly renders a lane-change decision.
}\label{fig:overview}\vspace{-0.1cm}
	\end{figure}

	Our work is inspired by recent successes that use a Deep Convolutional Neural
Network (DCNN) to control autonomous vehicles with end-to-end learning
fashion~\cite{Chen2015,Bojarski2016}. In~\cite{Bojarski2016}, the authors trained
a DCNN that directly maps an input image into a steer angle value. Chen~\etal\cite{Chen2015}~
aimed to obtain structured outputs of the driving-relevant objects from an image input rather than directly controlling a car. Our approach lies somewhere between the two algorithms, since our model produces a single output that is used as an aid to the final decision making process. In this work, we expect that a DCNN can learn valid image features so that it classifies the 
occupancy status of the lanes. Eventually, the proposed framework helps human drivers and autonomous 
vehicles avoid lane-change crashes.

	In this paper, we aim to develop an end-to-end learning framework that assists 
safe lane-change decision. Instead of object detection or tracking approaches, we formulate an image
classification problem that determines the status of adjacent lanes: \B~or
\F~(see Fig.~\ref{fig:overview}). Two cameras with a wide angle lens are installed at
the exterior of our research vehicle to acquire rear side view images. To
train and test Safe Lane-Change decision Aid Network (SLCAN), we collected and annotated
77,273 images. For an efficient SLCAN training, we annotate the images according to
whether the ego-vehicle can move on the corresponding space. The experimental results
on driving videos show that SLCAN classifies occupancy status of the lanes with an accuracy over 96.98\%.
   
	The main contributions of the paper are 
	\begin{itemize}
	\item A novel end-to-end learning system for lane-change is proposed that requires
	 no intermediate stages such as driving-relevant objects perception and risk prediction; 
	\item A new dataset of the rear side view images is collected and annotated to
	 train the proposed system.
	\end{itemize}

	The rest of paper is organized as follows. In Section~\ref{sec:slcan}, we propose
an end-to-end learning framework for safe lane-change decision aid. 
Section~\ref{sec:experiments} provides extensive experimental results. Finally, we
conclude this paper in Section~\ref{sec:conclusions}.

	\begin{figure}[t]
		\centering      
		\subfigure{\includegraphics[width=8.5cm]{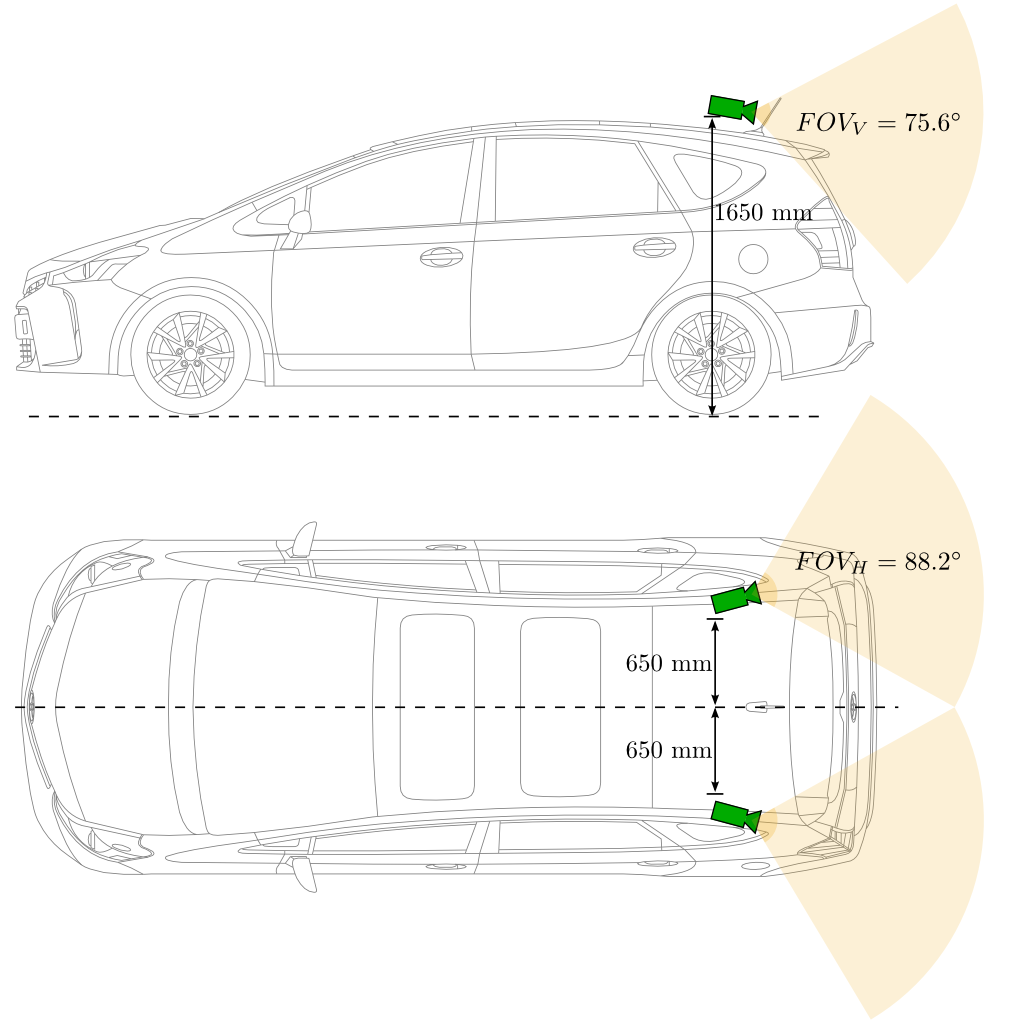}}
      	\caption{Our research vehicle senses rear side space with cameras mounted at
the both sides.}\label{fig:vehicle}\vspace{-0.3cm}
   \end{figure}

	\begin{figure}[t]
		\centering      
		\subfigure{\includegraphics[width=2.75cm]{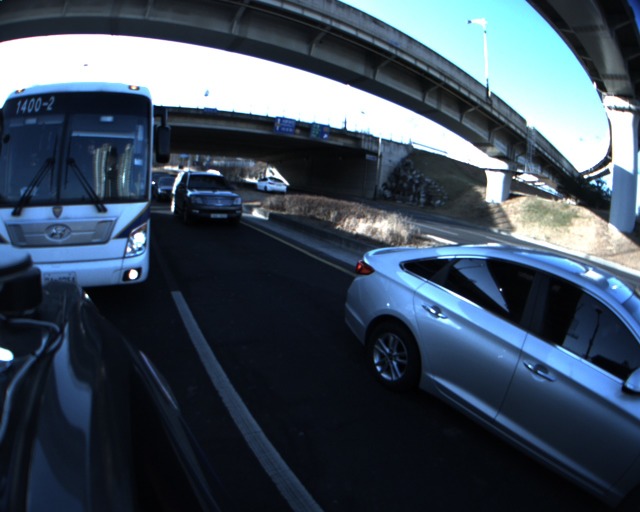}}%
		\setcounter{subfigure}{0}
		\subfigure[{\tt BLOCKED}]{\includegraphics[width=2.75cm]{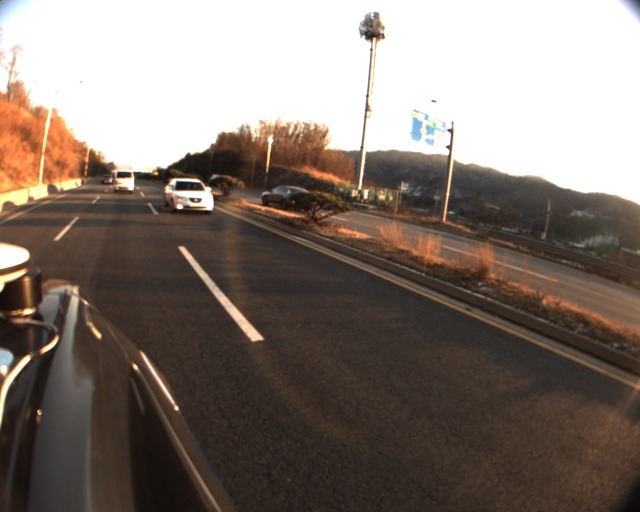}}%
		\subfigure{\includegraphics[width=2.75cm]{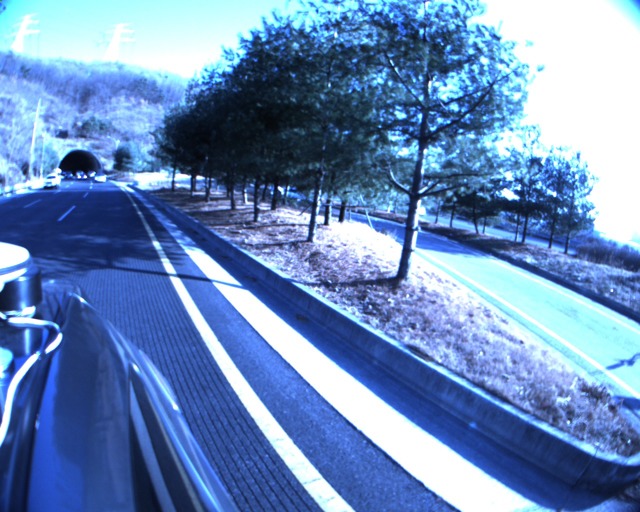}}\\
		\subfigure{\includegraphics[width=2.75cm]{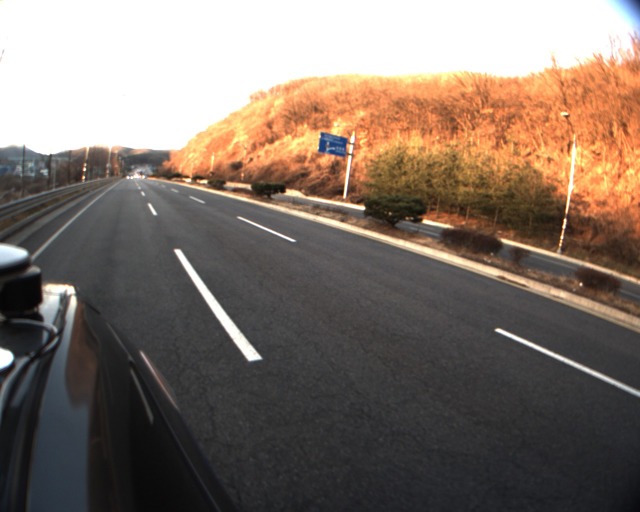}}%
		\setcounter{subfigure}{1}
		\subfigure[{\tt FREE}]{\includegraphics[width=2.75cm]{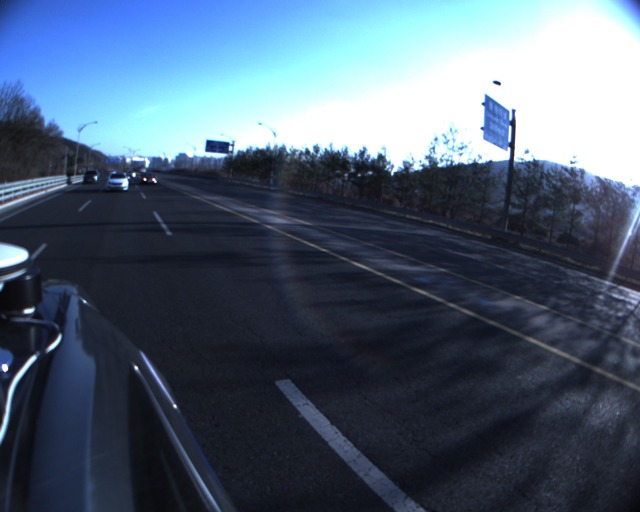}}%
		\subfigure{\includegraphics[width=2.75cm]{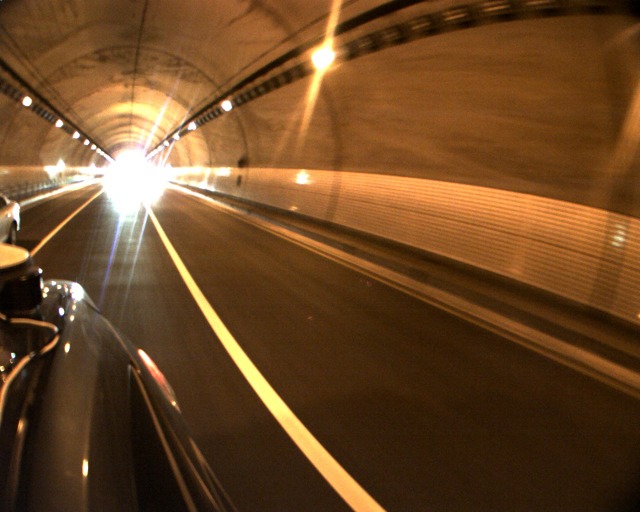}}\\
      
		\subfigure{\includegraphics[width=2.75cm]{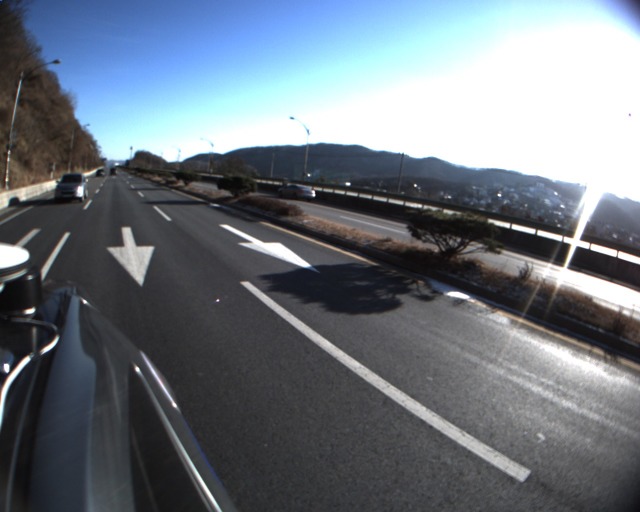}}%
		\setcounter{subfigure}{2}
		\subfigure[{\tt UNDEFINED}]{\includegraphics[width=2.75cm]{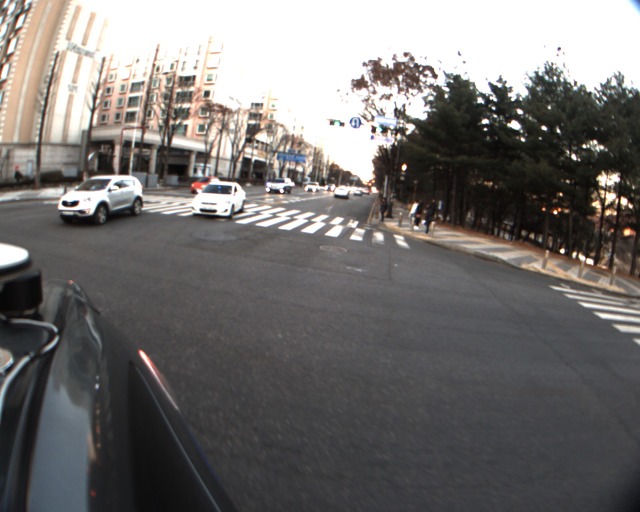}}%
		\subfigure{\includegraphics[width=2.75cm]{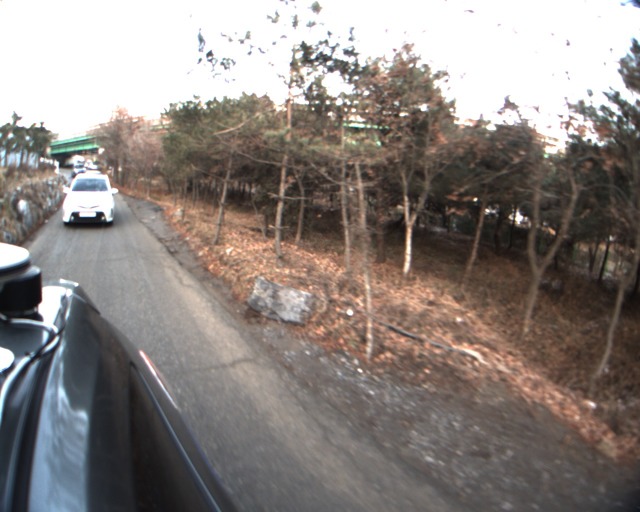}}
      
		\caption{Examples of the annotated images for left rear side view: (a) \B, (b) \F, and (c) \ud}\label{fig:examples}
   \end{figure}
   
\section{Safe Lane-Change decision Aid Network}
\label{sec:slcan}

	Our goal is to design a DCNN that can tell whether there is enough room for lane-%
change at the moment.  Unlike the previous approaches~\cite{schubert2010,Pellkofer2002,Schlechtrienmen2014,Damerow16,Kesting99}, 
we integrate perception and decision making process into a single image classification 
process. Intuitively, our approach is more like human driver's behavior in that a human
driver's decision is based on a glance of very short time instead of systematic analysis
of surrounding traffic situation.  We define the image based lane-change decision problem
as follows.
	
	Let $f:I\mapsto$\{\B, \F\}$\in\mathbb{A}$ be a function that classifies the given
image $I$ onto occupancy status of the corresponding lane. To our best knowledge, we
have no dataset for this specific task. We thus create lane-change decision aid
dataset which are pairs of image and lane occupancy status $\mathfrak{D}=\{(I_k,
a_k)\}_{k=1,\dots K}$, where $a\in\mathbb{A}$ and $K$ denotes the number of images in the dataset. We feed the labelled images to a DCNN in order to learn a function that
decides the initiation of the lane-change. For training, softmax loss is employed to
quantify the prediction quality compared to groundtruth. With a trained DCNN model, we
obtain values which correspond to likelihood probabilities of the current occupancy
status for the given image $I$. Finally we classify the lane status of the input image
as follows:
	\[
		f(I) = \argmax_{a\in\mathbb{A}}P(a\mid I).
	\]	

	\subsection{Collected Data Annotation}
	SLCAN accepts rear-side view images that are acquired by cameras mounted on 
the left and right side of the vehicle.  To train SLCAN, we need to annotate each 
acquired image with one of three labels: \B,~\F, and \ud.  An image is tagged as \B~if 
the adjacent lane is occupied by other vehicles or road structures.  On the contrary, 
an image is tagged as \F~if the lane is physically clear enough to initiate lane-change 
in a few seconds.  Here, note that physical clearance overrides the traffic regulations.  
That is, even in the case that the lane-change is not allowed by a regulation 
(\eg changing lane in a tunnel, or even crossing yellow center line), we annotate such 
images as \F~ if the ego-vehicle can physically move to that lane space. The rationale 
is that SLCAN is not a final decision maker but an aid module so it should provide as 
much information as possible to human driver or main controller of the autonomous vehicle. 
In the case that we cannot determine whether an image is \B~or \F, we annotate the image 
as \ud; for examples, the ego-vehicle is already in the middle of lane-change motion,
making a turn at an intersection, or under any other ambiguous situations. To minimize 
ambiguity in classification, we exclude \ud~labelled images in the training dataset. 
Fig.~\ref{fig:examples} shows examples of the annotation results. Moreover, in this work, 
we use images of typical roadways only; therefore, images of narrow paths or unpaved 
roads are discarded.
	
	We summarize our annotation criteria as follows:
	\begin{itemize}\it
		\item {\tt BLOCKED} if the ego-vehicle cannot physically move to the corresponding space;
		\item {\tt FREE} if the ego-vehicle can move to the corresponding space even if such action may violate the traffic regulations;
		\item {\tt UNDEFINED} for an ambiguous situation and any other unsusual scenes.
	\end{itemize}

	A human driver has his/her own decision rule to draw a function $f(\cdot)$ 
based on individual driving technique and experience. To construct reliable 
groundtruth $g(\cdot)$, for a given image, we accept the annotation result when all 
annotation workers agree on; otherwise, the image is tagged as \ud. More formally,

	\begin{equation}
		g(I) = \bigwedge_{n=1}^N{f}_n(I), 
	\end{equation}
where $f_n(I)$ denotes the annotated image by $n$-th worker and $\wedge$ is defined as
	\[
		a \wedge b = \left\{
			\begin{array}{ll}
				\texttt{BLOCKED},& \mbox{if $a$ and $b$ are both \texttt{BLOCKED}}\\
				\texttt{FREE},& \mbox{if $a$ and $b$ are both \texttt{FREE}}\\
				\texttt{UNDEFINED},& \mbox{otherwise.}\\				
			\end{array}
			\right.
	\]
For all images, at least three workers annotate with respect to the lane status.
	\subsection{Network Architecture}
    We use the VGG 16-layer architecture~\cite{Simonyan2014}, with the original 1000-%
way final classification layer replaced by a 2-way classifier. We take the model pre-%
trained on ILSVRC 2012 dataset~\cite{Deng2009} and fine-tune it with the dataset we
collected to classify the rear view image into two classes: \B~and
\F. In the following section, we give a detailed description of the experiment setup
and training procedure.
	
	\begin{figure}[t]
		\centering      
		\subfigure[]{\includegraphics[width=4.2cm]{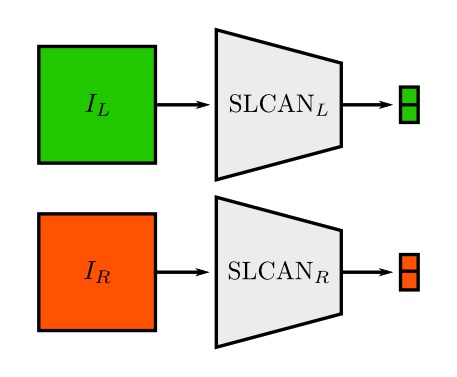}}~
		\subfigure[]{\includegraphics[width=4.2cm]{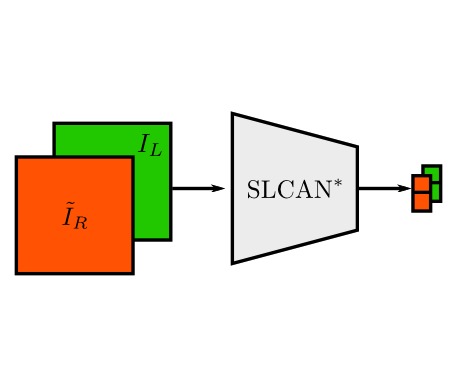}}
		\caption{Illustration of SLCAN architecture configurations: (a) two independent
networks that separately train left rear view image $I_L$ and right rear view image $I_R$, 
and (b) unified single network that takes left rear view image $I_L$ and flipped right 
rear view image $\tilde{I}_R$. When deployed on a GPU, configuration of (b) can process
two input images in parallel, we save memory space without additional processing time.
}\label{fig:networks}\vspace{-0.3cm}
	\end{figure}

	\begin{table}[tbp]
	\centering
	\setlength\tabcolsep{2pt}
	\caption{Data configuration and classification results$^a$}\label{t:two}
	\begin{tabular}{c c c c c c c}
	\hline
	\multirow{2}{*}{Exp.} & \multirow{2}{*}{Model} & \multirow{2}{*}{Accuracy} & & \multicolumn{3}{c}{Dataset}\\
	\cline{5-7}
	& & & & {\tt BLOCKED} & {\tt FREE} & Total \\ \hline\hline
	\multirow{6}{*}{\bf E1} & \multirow{3}{*}{${\rm SLCAN}_L$} & \multirow{2}{*}{99.77\%} & Training & 18,390 & 15,913 & 34,303 \\
	& & \multirow{2}{*}{@Val} & Validation & 2,043 & 1,768 & 3,811 \\
	& & & Test &- & - & - \\
	\cline{2-7}
	& \multirow{3}{*}{${\rm SLCAN}_R$} & \multirow{2}{*}{99.70\%} & Training & 20,512 & 14,732 & 35,244 \\
	& & \multirow{2}{*}{@Val} & Validation & 2,279 & 1,636 & 3,915  \\
	& & & Test & - & - & -\\
	\hline
	\multirow{3}{*}{\bf E2} & \multirow{3}{*}{$\rm SLCAN^*$} & \multirow{2}{*}{99.90\%} & Training & 38,902 & 30,645 & 69,547 \\
	& & \multirow{2}{*}{@Val} & Validation & 4,322 & 3,404 & 7,726 \\
	& & & Test & - & - & - \\
	\hline
	\multirow{3}{*}{\bf E3} & \multirow{3}{*}{$\rm SLCAN^*$} & \multirow{2}{*}{96.98\%} & Training & 34,092 & 27,284 & 61,376 \\
	& & \multirow{2}{*}{@Test} & Validation & 3,787 & 3,031 & 6,818 \\
	& & & Test & 5,345 & 3,734 & 9,079 \\
	\hline
	\multicolumn{7}{l}{$^a$In all cases, the networks were trained up to 5000 epochs.}\\
	\end{tabular}
	\end{table}

\section{Experiments}
\label{sec:experiments}

	\subsection{Data Description}
	We mounted and synchronized two Point Grey Blackfly cameras on the left and right rooftop of the
vehicle so that both cameras sense each rear side view space. The camera has 88.2 degrees
horizontal and 75.6 degrees vertical field of view (FOV) with lens of 3.5 mm focal
length. Images of 1280$\times$1024 pixel resolution are acquired at the rate of 10 frames per second.

	We drove our research vehicle on highway and urban roadway to acquire various 
scenes and collected a total of 100,088 images (50,044 pairs of left and right images).
The images of the dataset consist of various 
road types (highway, intersection, merge, fork, tunnel, and \etc), 
traffic conditions (from free flow to congestion), 
types of vehicles (car, truck, bus, van, motorcycle, and \etc),
and on-road objects and markings (barrier, curb, fence, cone, crosswalk, no stopping zone, and \etc).

	After annotation work, we set aside all images tagged as {\tt UNDEFINED} and
finally have 38,114 left rear view images (20,433 {\tt BLOCKED} and
17,681 {\tt FREE}) and 39,159 right rear view images (22,791 {\tt BLOCKED} and
16,368 {\tt FREE}).

   \subsection{Configurations}

    We conducted the following three experiments with different network setups and dataset splits:

    \begin{enumerate}[\bf E1.]

    \item{\bf Two independent networks for left and right rear view}

    First, we separately trained two DCNNs for left and right view images, as
shown in Fig.~\ref{fig:networks} (a). Since the two camera views are not symmetric 
due to the road configuration (\eg centerline is always observed 
in the left rear view images), we supposed that each rear view would need to learn 
its own model. We split the dataset into 90\% training set and 10\% validation set 
(see Table~\ref{t:two}).

    \item{\bf Single network for both cameras}

	As depicted in Fig.~\ref{fig:networks} (b), we also trained an unified 
network for both rear view images, where right view images are horizontally flipped
to match left rear view. Training and validation sets were formed by combining the 
corresponding dataset from both rear view images. The advantage of using a single 
network is that it can simultaneously process left rear image and flipped right rear 
image in a batch on a GPU. It requires no additional processing time while only half 
of the memory space is used comparing to two independent networks architecture.

    \item{\bf Single network for both cameras, trained on highway images and tested on urban road images}

    To see how well our model generalizes to datasets outside the training set, 
we also experimented training the single network model on highway portion of the dataset 
and testing it on urban road portion of the dataset. The number of samples for each dataset 
is given in Table~\ref{t:two}.

    \end{enumerate}

   \subsection{Training}

\begin{figure*}[t]
		\centering      
		\subfigure[Left rear view image $I_L$]{\includegraphics[width=5.60cm]{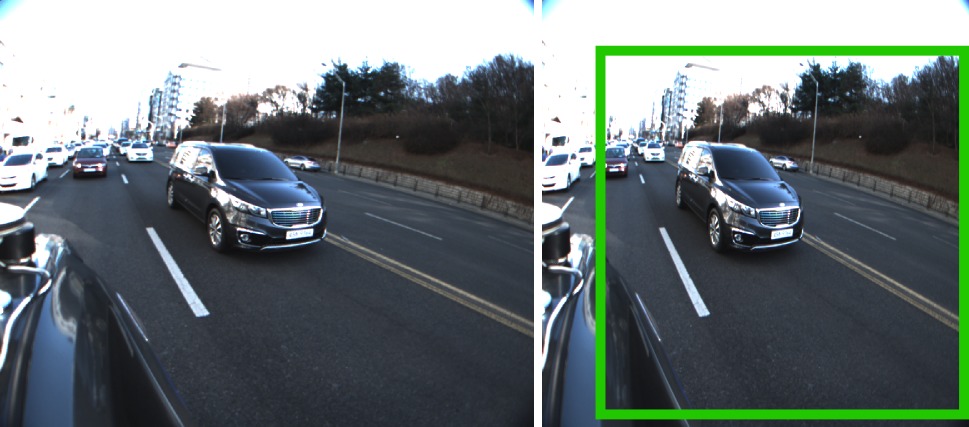}}~
		\subfigure[Right rear view image $I_R$]{\includegraphics[width=5.60cm]{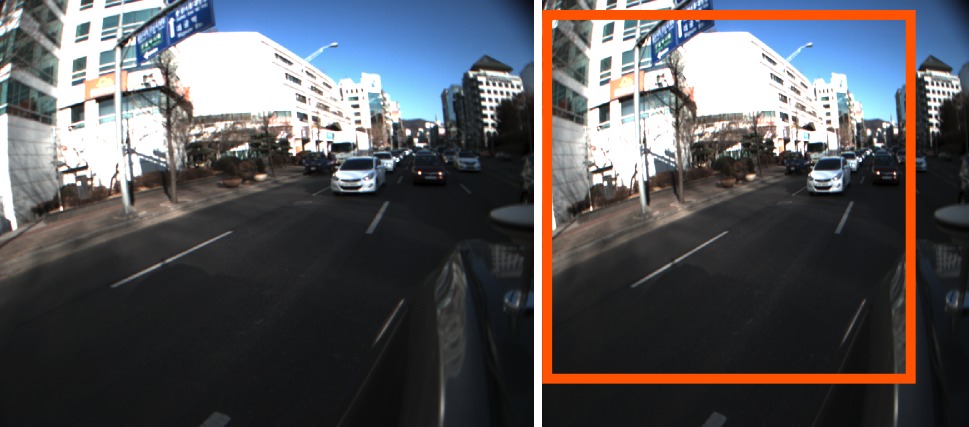}}~
		\subfigure[Flipped right rear view image $\tilde{I}_R$]{\includegraphics[width=5.60cm]{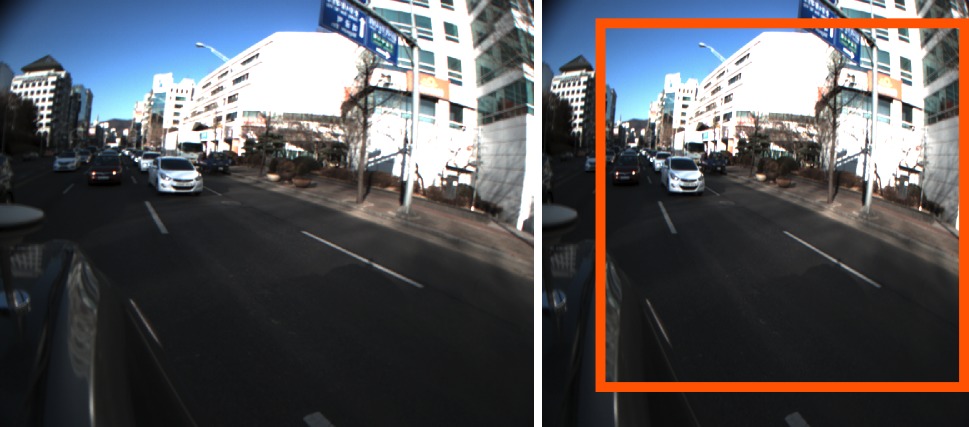}}\vspace{-0.1cm}
		\caption{Illustration of resizing and cropping process for SLCAN input}\label{fig:roi}
	\end{figure*}

    Unlike images in general image classification task such as ImageNet, pixels 
in our road view images have unequal influence on the classification results 
depending on their location. For example, pixels near leftmost column in left camera 
view corresponds to the ego-lane and thus can be ignored, while those near rightmost 
column must not be discarded by cropping as in ImageNet training because it can contain 
the tail end of the vehicle in an adjacent lane. To preserve such spatial variance of 
our data, we resize the input images to 256$\times$256, then a 224$\times$224 patch is 
cropped at fixed horizontal offset (32 pixels for left camera, 0 pixels for right camera), 
and random vertical offset (at training time) or vertical center (at test time). 
For the same reason, we also omit random horizontal flip which is normally employed 
for data augmentation purpose. For the single network experiments, images from right 
camera are horizontally flipped. Please see Fig.~\ref{fig:roi} for an illustration.

    For all experiments, we fine-tuned the ImageNet-pretrained VGG-16 network with 
a minibatch size of 64 and learning rate of 0.001 until convergence. We used 
Caffe~\cite{jia2014caffe} to implement the experiments.

	\begin{figure*}[!t]
	\begin{center}
	\footnotesize
	\begin{tabular}{*{4}{>{\centering\arraybackslash} m{4.cm}}}
		\includegraphics[width=4.25cm]{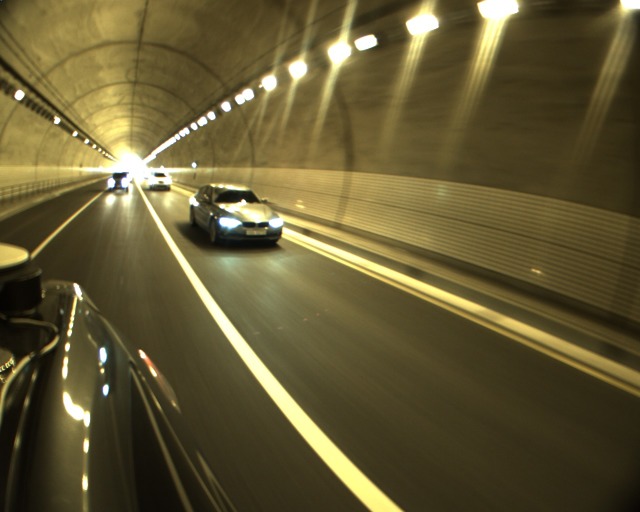} &
		\includegraphics[width=4.25cm]{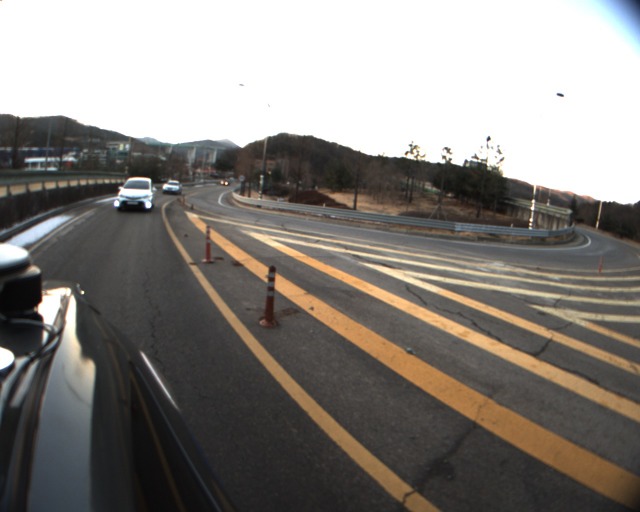} &
		\includegraphics[width=4.25cm]{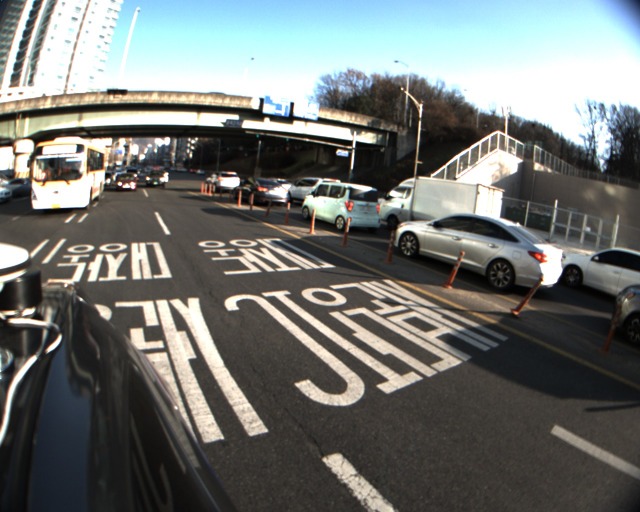} &
		\includegraphics[width=4.25cm]{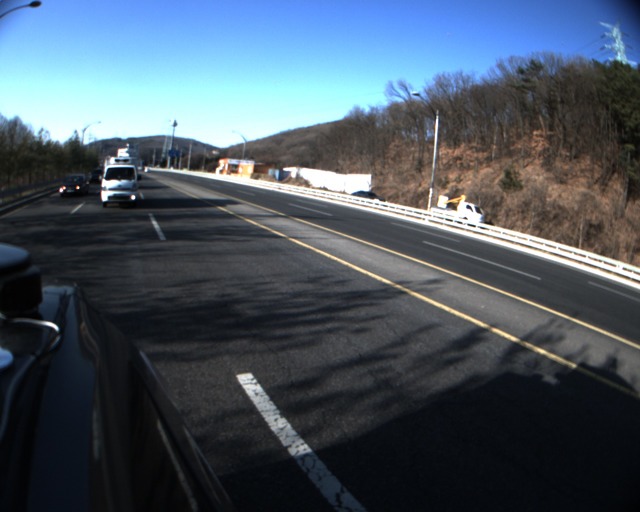}\\
		$P(\B\mid I) = \color{blue}{1.00}$ &
		$P(\B\mid I) = \color{blue}{1.00}$ &		
		$P(\F\mid I) = \color{blue}{1.00}$ &	
		$P(\B\mid I) = \color{red}{0.73}$ \\	 
		\multicolumn{4}{c}{(a) results of ${\rm SLCAN}_L$} \\
		\includegraphics[width=4.25cm]{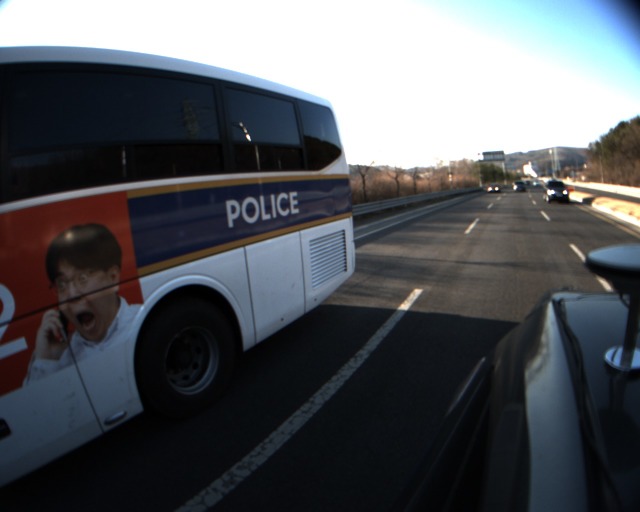} &
		\includegraphics[width=4.25cm]{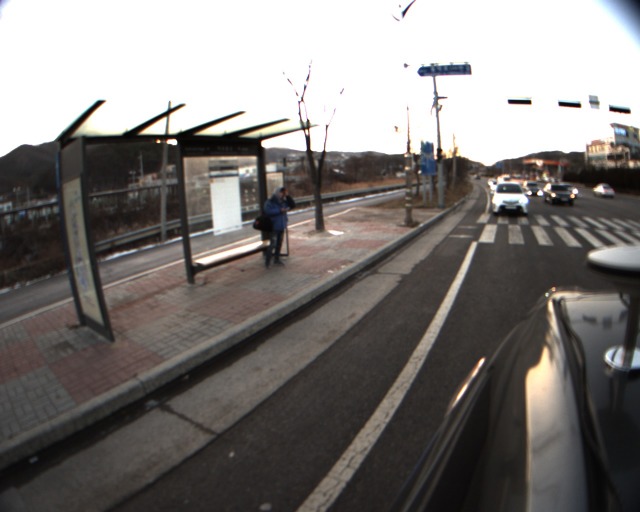} &
		\includegraphics[width=4.25cm]{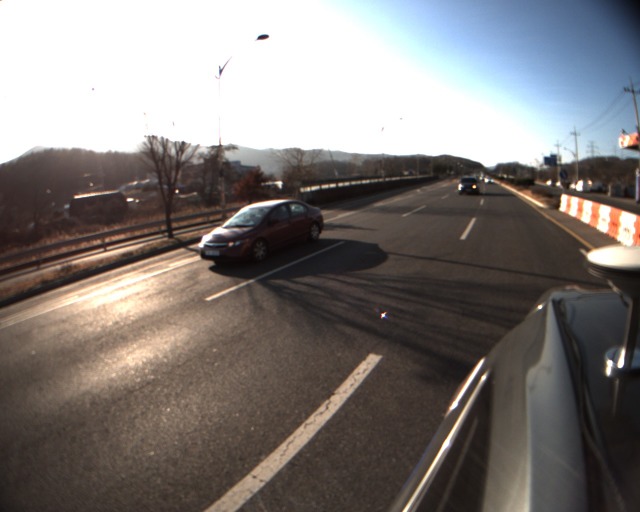} &
		\includegraphics[width=4.25cm]{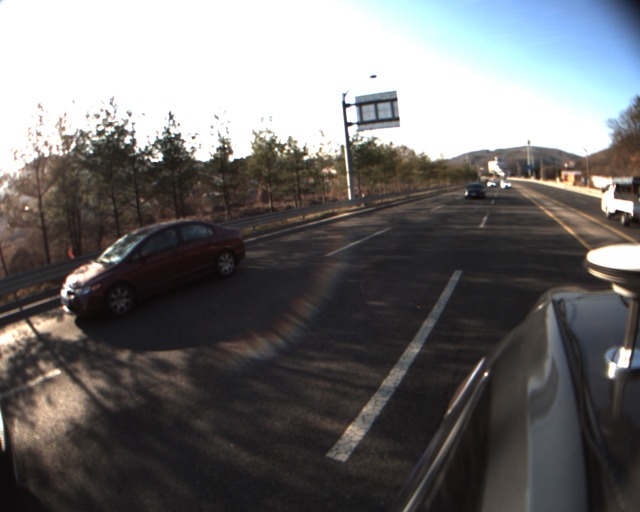} \\
		$P(\B\mid I) = \color{blue}{1.00}$ &	
		$P(\B\mid I) = \color{blue}{1.00}$ &	
		$P(\F\mid I) = \color{blue}{1.00}$ &	
  		$P(\F\mid I) = \color{red}{0.82}$ \\
		\multicolumn{4}{c}{(b) results of ${\rm SLCAN}_R$} \\
		\includegraphics[width=4.25cm]{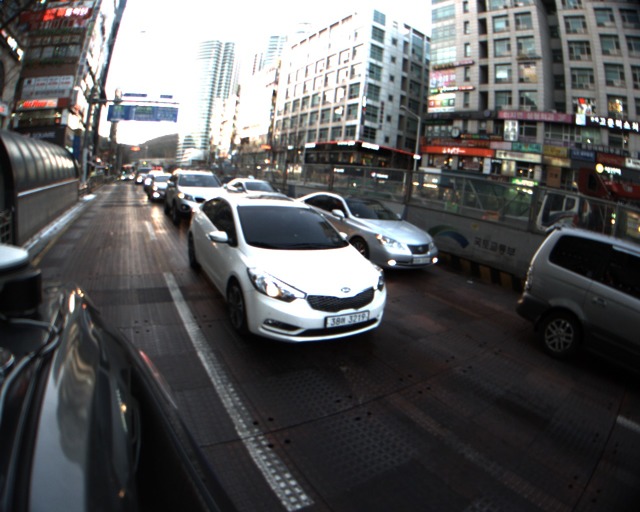} &
		\includegraphics[width=4.25cm]{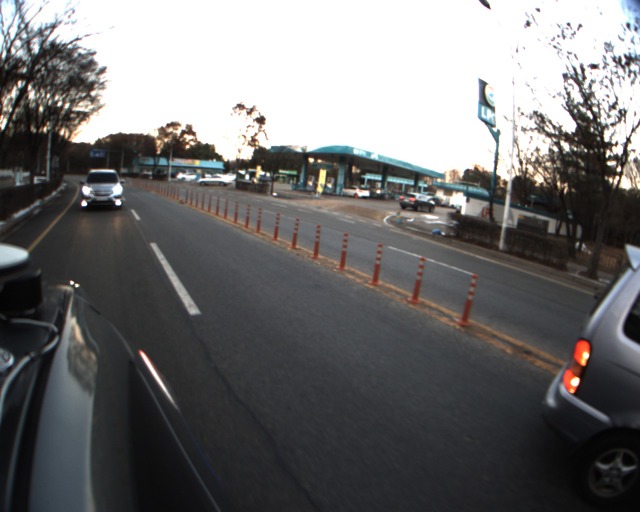} &
		\includegraphics[width=4.25cm]{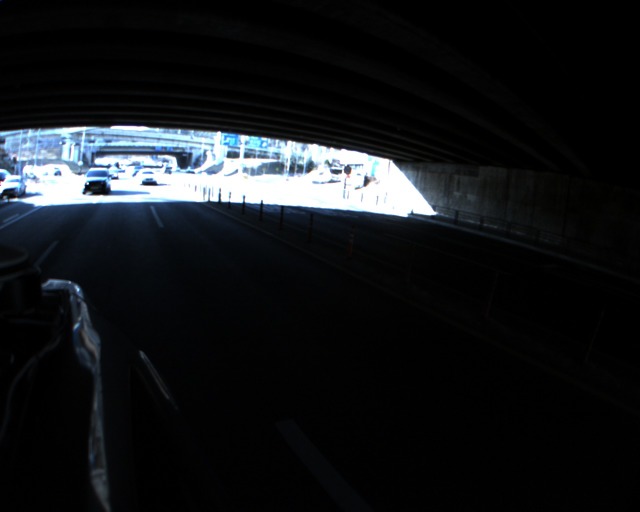} &
		\includegraphics[width=4.25cm]{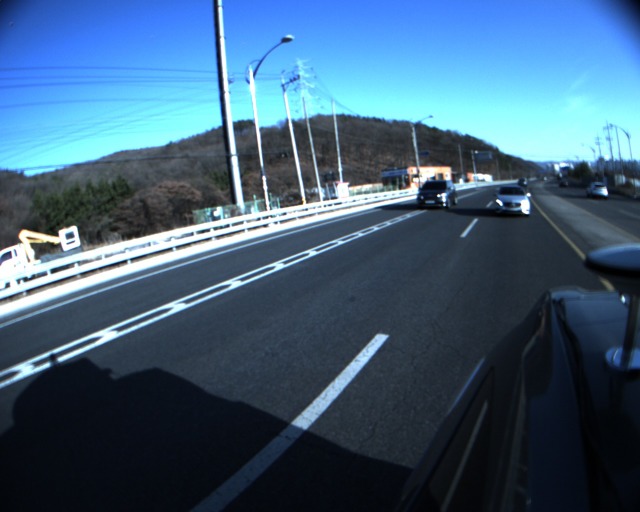} \\
		$P(\B\mid I) = \color{blue}{1.00}$ &	
		$P(\B\mid I) = \color{blue}{1.00}$ &	
		$P(\F\mid I) = \color{blue}{1.00}$ &	
		$P(\F\mid I) = \color{red}{1.00}$ \\
		\multicolumn{4}{c}{(c) results of ${\rm SLCAN}^*$} \\  	  	  	  
	\end{tabular}
	\end{center}\vspace{-0.25cm}
	\caption{Experimental results for ${\rm SLCAN}_L$, ${\rm SLCAN}_R$, and ${\rm SLCAN}^*$. 
Colored numbers indicate the probability value returned by DCNN, where blue and red colors
denote correct and incorrect classification, respectively. Most of the incorrect 
classifications (on the rightmost column) actually correspond to situations that are 
ambiguous or difficult even for humans. (a) The shadows cast by a truck and trees lead to misclassification. (b) The vehicle on the right lane is too far behind to
precisely determine the occupancy status of the lane. (c) A vehicle is merging into 
the right lane.}\label{fig:results}\vspace{-0.45cm}
	\end{figure*}  

	\subsection{Results}
         
    We found that a single network trained with left rear view images and horizontally
flipped right rear view images performs just as well as networks separately trained for
each rear view image. As shown in Table~\ref{t:two}, we obtained nearly identical validation 
accuracies for both approaches. Some examples of correct and incorrect classification are 
shown in Fig.~\ref{fig:results}. Incorrectly classified samples include borderline cases 
with adjacent lane vehicle appearing very small at a far distance, and frame captured 
under uneven lighting condition.

\begin{figure}[t]
	\centering
	\footnotesize
	\begin{tabular}{*{2}{>{\centering\arraybackslash} m{3.8cm}}}
		\includegraphics[width=4.0cm]{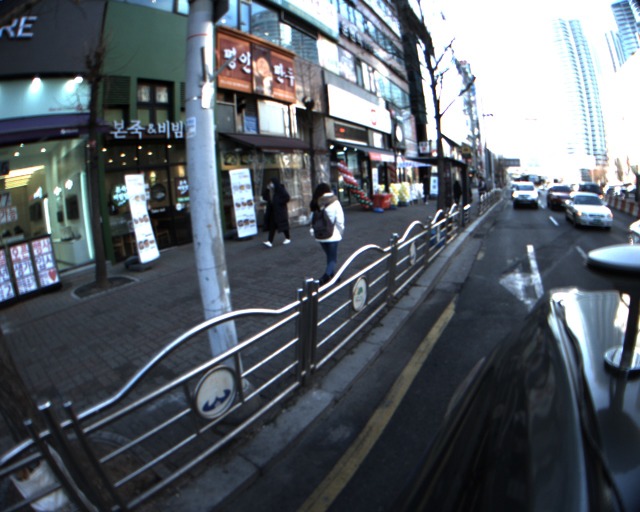} &
		\includegraphics[width=4.0cm]{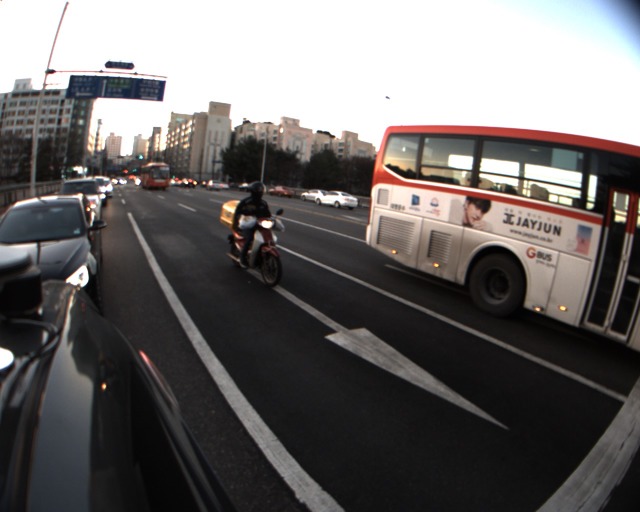} \\
		(a) $P(\B\mid I) = \color{blue}{1.00}$ &	
		(b) $P(\B\mid I) = \color{blue}{1.00}$ \\
	
		\includegraphics[width=4.0cm]{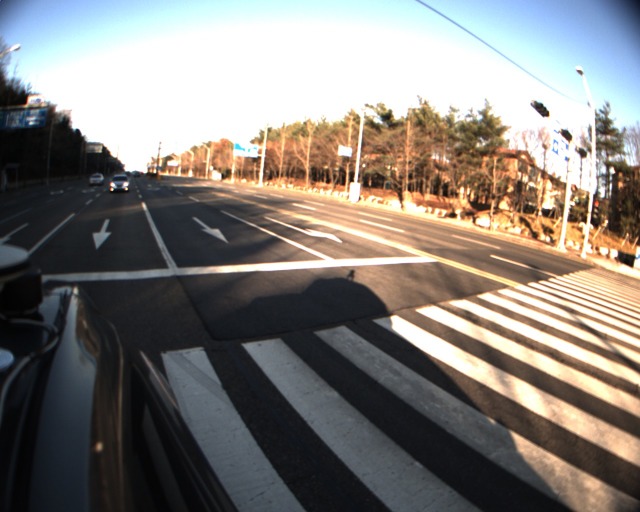} &
		\includegraphics[width=4.0cm]{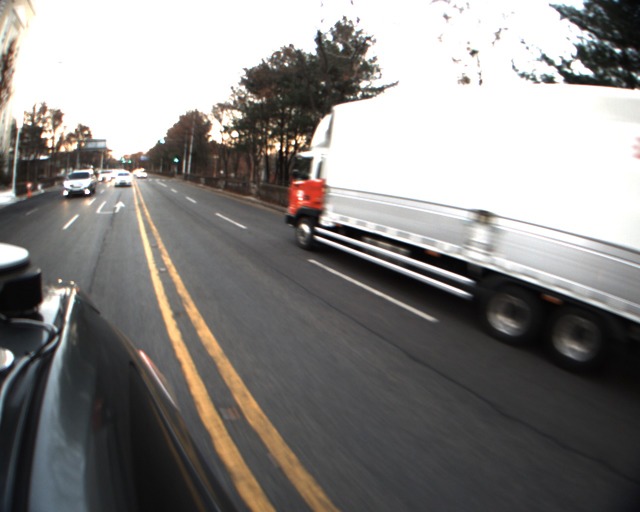} \\
		(c) $P(\B\mid I) = \color{red}{0.59}$ &	
		(d) $P(\B\mid I) = \color{red}{0.90}$ \\	  
	\end{tabular}
	\caption{Experimental results of correct and incorrect classification with 
$\rm SLCAN^*$, where the test images are new to the system. 
(a)--(b): Images with fences and motorcycle, which do not appear in training data, are correctly classified. 
(c)--(d): $\rm SLCAN^*$ is confused to classify \B~status with unknown road markings and trailer. Note that the overall classification accuracy is 96.98\% and misclassification of this kind is very scarce. 
	}\label{fig:generalize}\vspace{-0.5cm}
  \end{figure}

    In Experiment 3, the accuracy slightly decreased by $\sim$3\%. As shown in 
Fig.~\ref{fig:generalize}, the model is able to correctly classify images containing 
some obstacles unseen in the training data such as fences or motorcycle, but makes 
wrong prediction for unseen road markings or large vehicles like a trailer. Still, 
it performs very well in general, indicating that our model successfully generalizes 
to unfamiliar scenes.

\begin{figure}[t]
	\centering
	\footnotesize
	\begin{tabular}{*{4}{>{\centering\arraybackslash} m{1.7cm}}}
		\includegraphics[width=2.0cm]{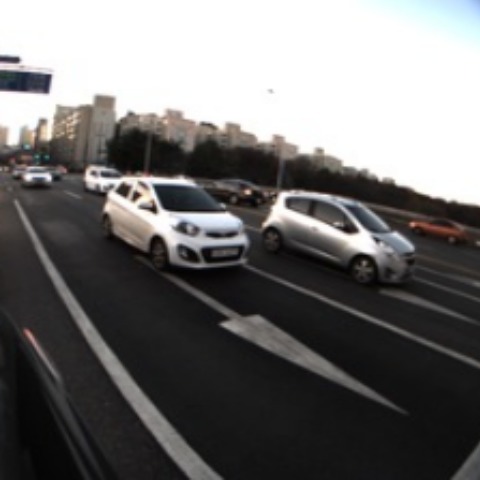} &
		\includegraphics[width=2.0cm]{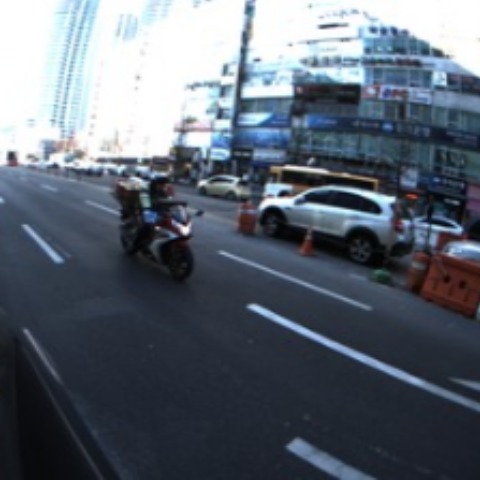} &
		\includegraphics[width=2.0cm]{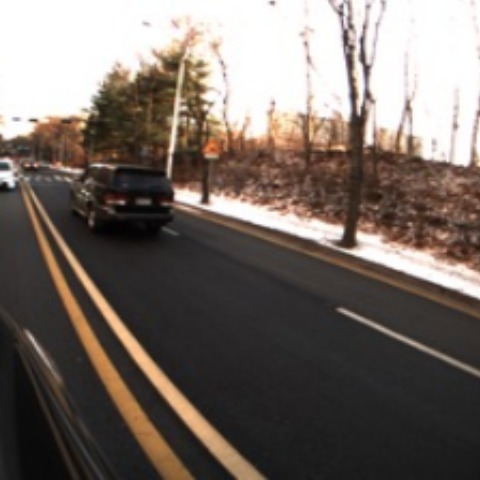} &
		\includegraphics[width=2.0cm]{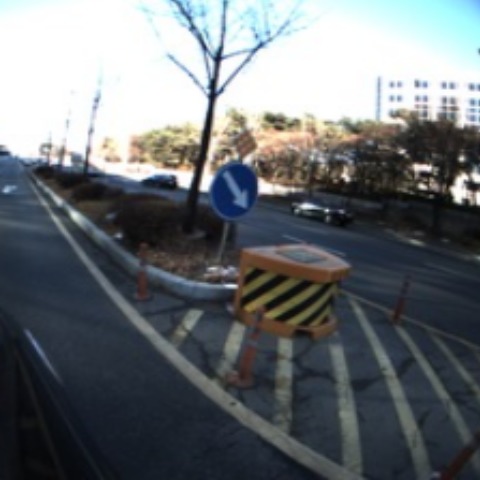} \\
		\includegraphics[width=2.0cm]{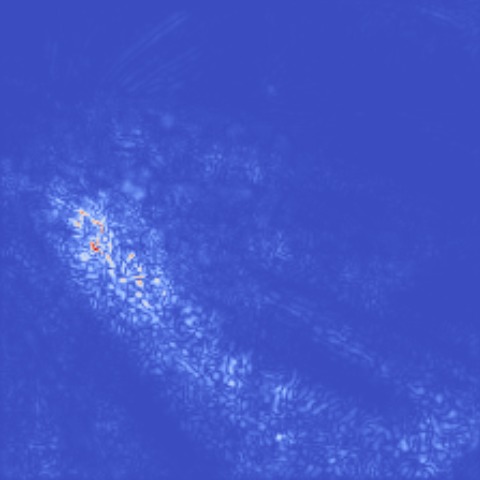} &
		\includegraphics[width=2.0cm]{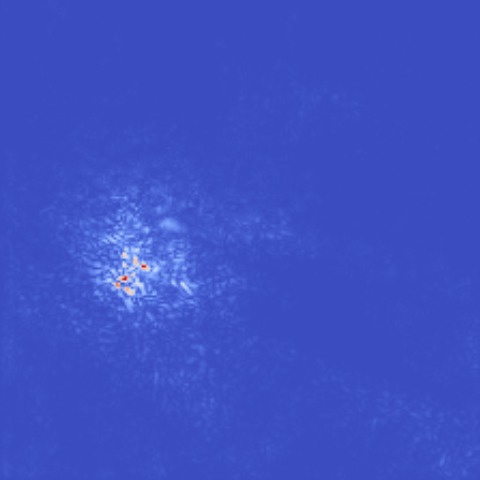} &
		\includegraphics[width=2.0cm]{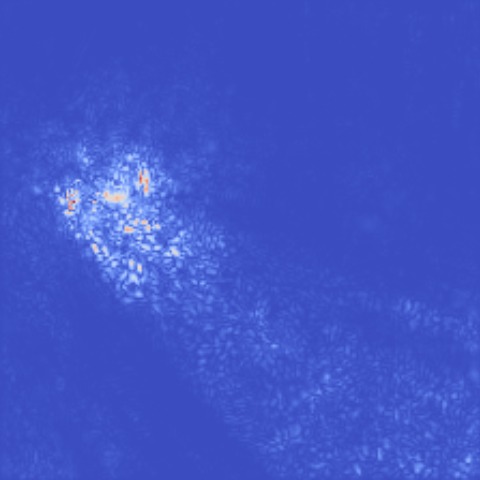} &
		\includegraphics[width=2.0cm]{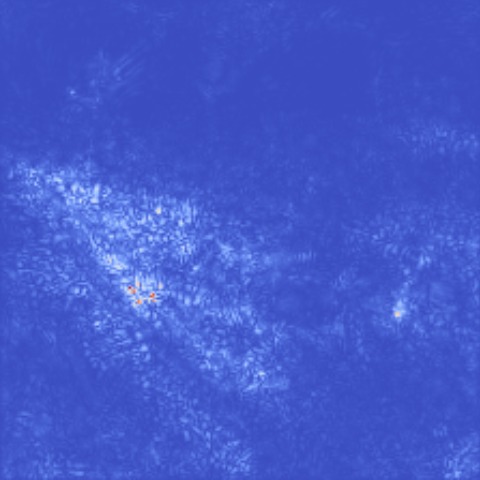} \\
		\multicolumn{4}{c}{(a) {\tt BLOCKED}} \\
		\includegraphics[width=2.0cm]{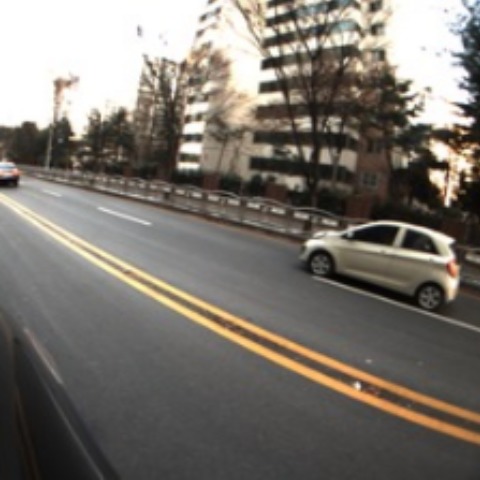} &
		\includegraphics[width=2.0cm]{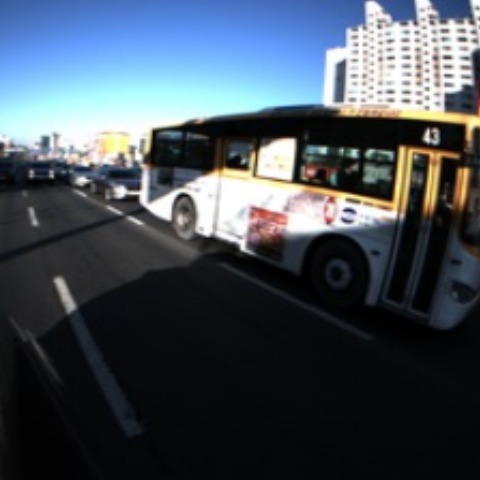} &
		\includegraphics[width=2.0cm]{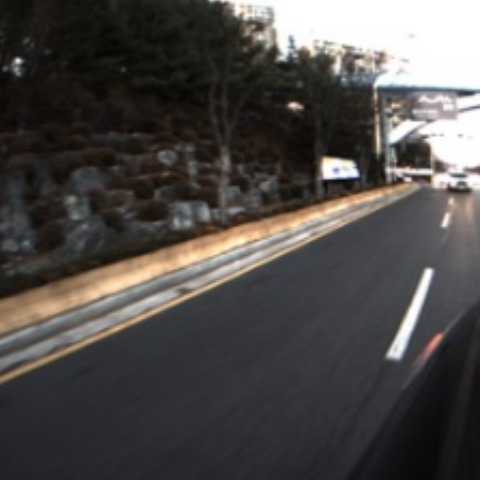} &
		\includegraphics[width=2.0cm]{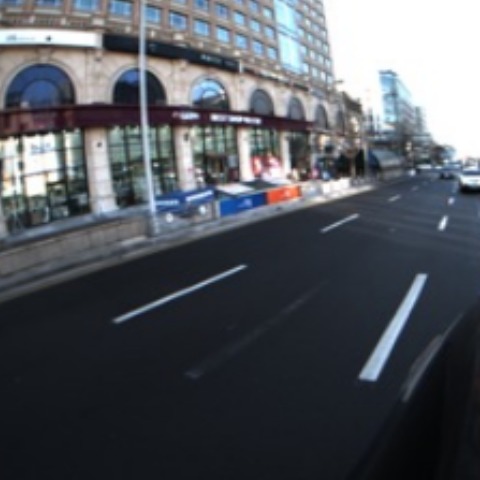} \\
		\includegraphics[width=2.0cm]{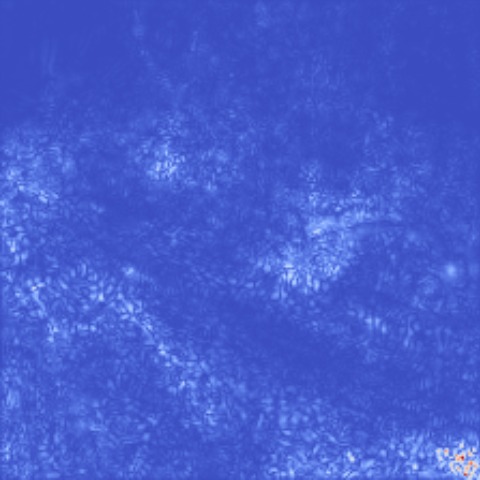} &
		\includegraphics[width=2.0cm]{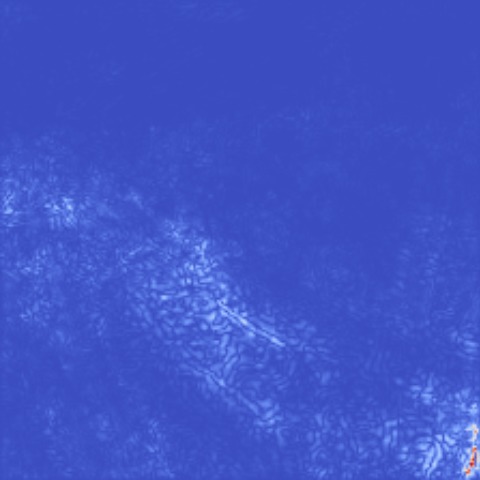} &
		\includegraphics[width=2.0cm]{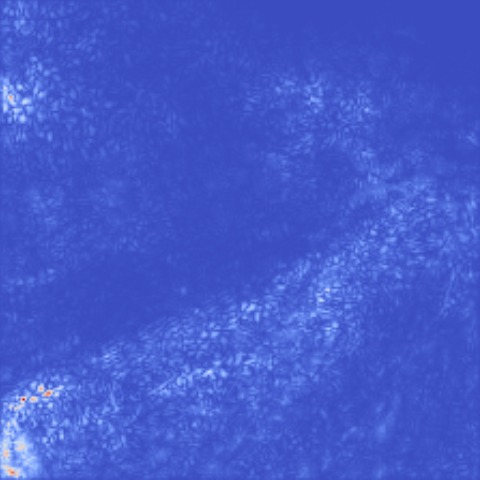} &
		\includegraphics[width=2.0cm]{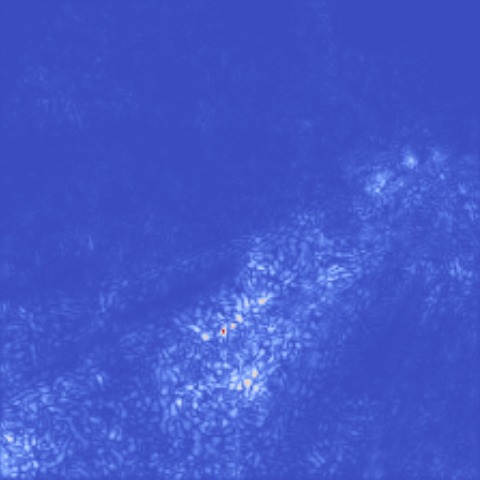} \\
		\multicolumn{4}{c}{(b) {\tt FREE}} \\
	\end{tabular}\vspace{-0.15cm}
	\caption{Saliency maps~\cite{Simonyan2014} that show where on the image SLCAN 
focuses on when it classifies an image. It is apparent that SLCAN focuses on 
blocking obstacles when they exist and on the road surface when the lane is free.}
	\label{fig:visualize}
	\end{figure}

    In Fig.~\ref{fig:visualize}, we show saliency maps~\cite{Simonyan2014a} that
visualize which part of the input image influences the classification results 
the most, by analyzing the magnitude of pixelwise gradients obtained by back-propagation. 
We can see that the network focuses on obstacles for \texttt{BLOCKED} images, and the 
road surface of the adjacent lane for \texttt{FREE} images.
    
\begin{figure}[th]
	\centering            
	\footnotesize
	\subfigure{\includegraphics[width=8.2cm]{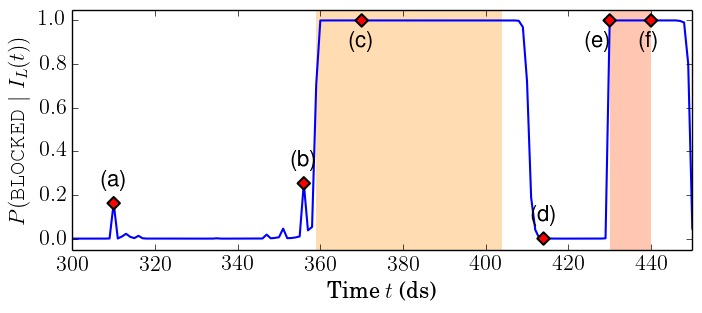}}\\
	\begin{tabular}{*{3}{>{\centering\arraybackslash} m{2.2cm}}}
		\includegraphics[width=2.3cm]{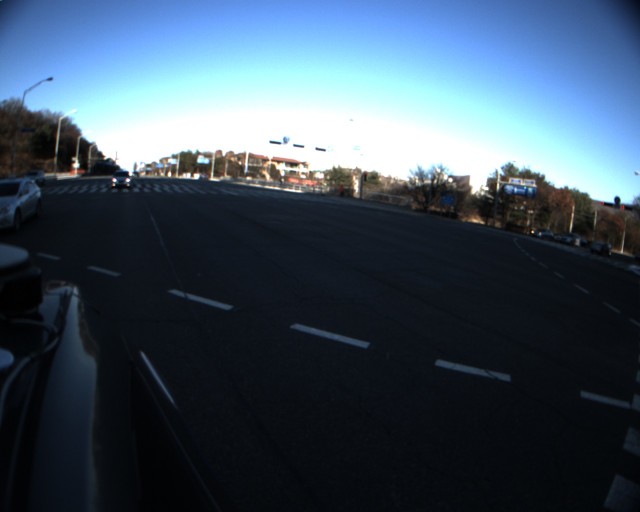} &
		\includegraphics[width=2.3cm]{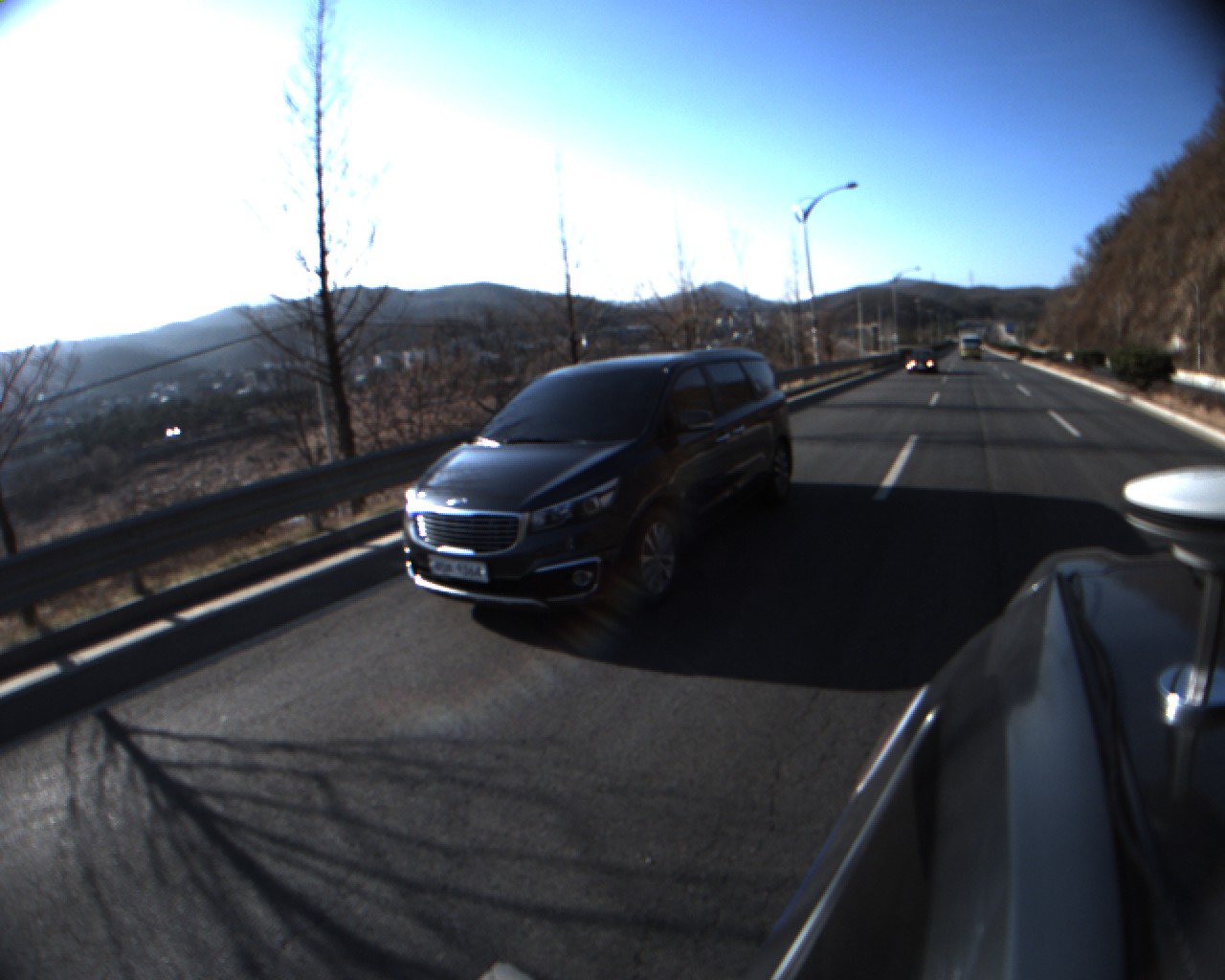} &
		\includegraphics[width=2.3cm]{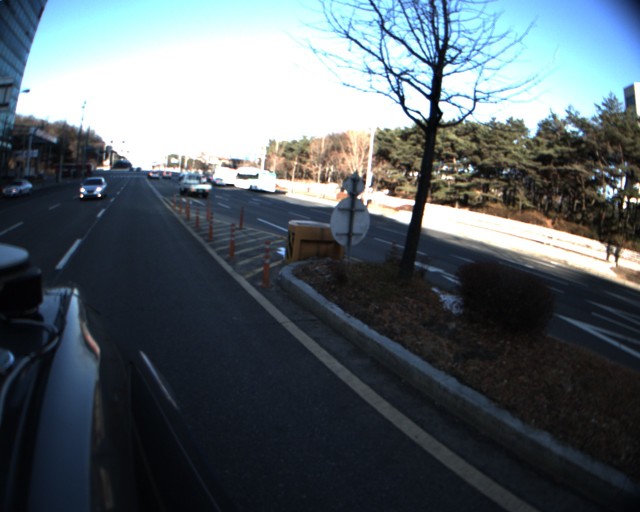} \\
		~(a) & ~(b) & ~(c) \\
		\includegraphics[width=2.3cm]{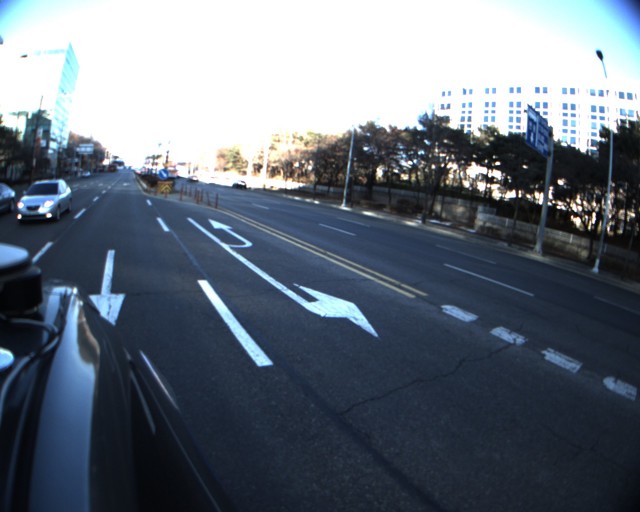} &
		\includegraphics[width=2.3cm]{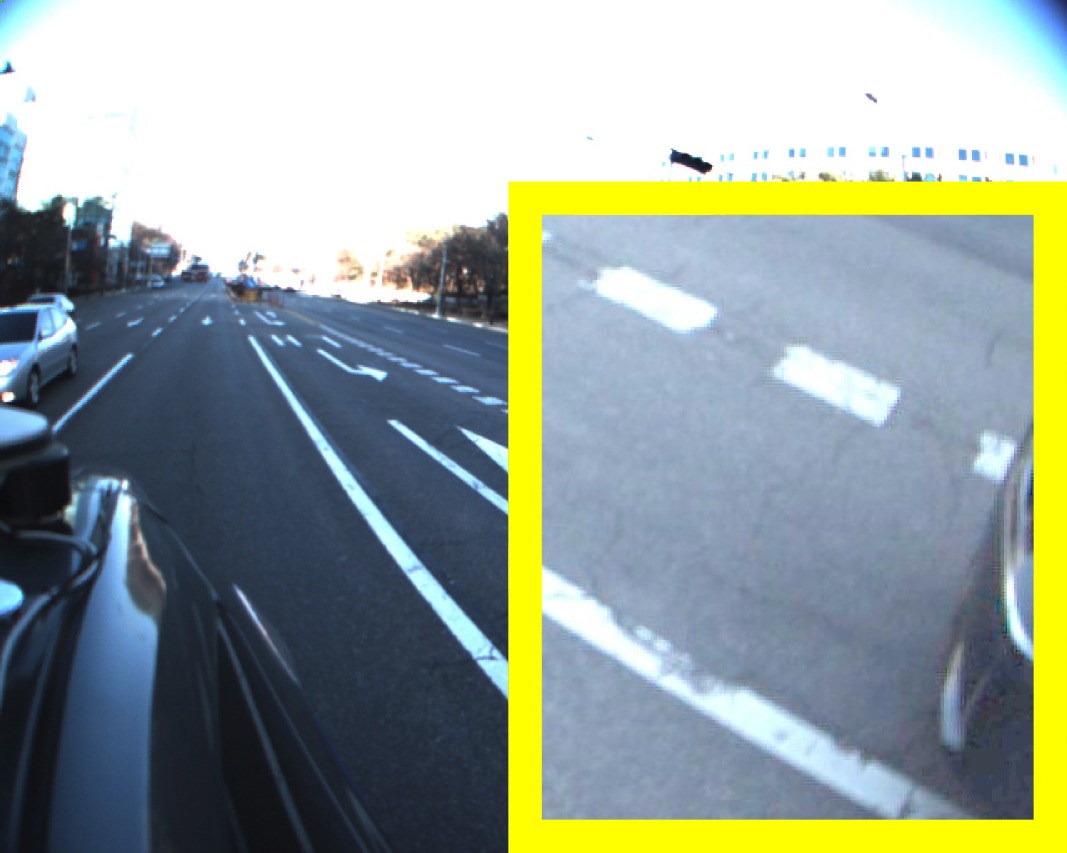} &
		\includegraphics[width=2.3cm]{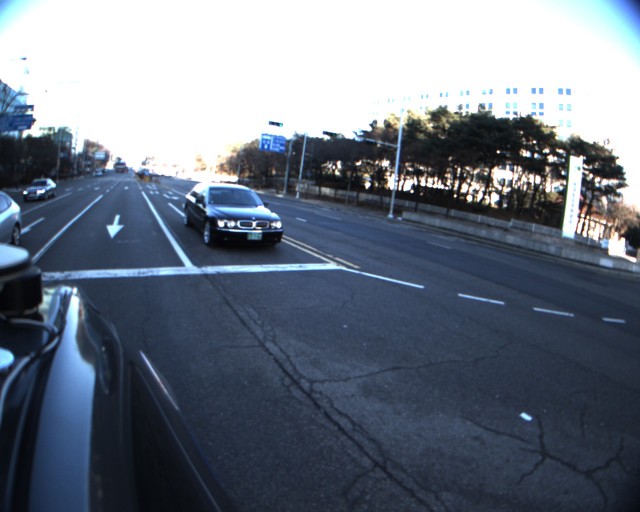} \\
		~(d) & ~(e) & ~(f) \\
	\end{tabular}  
	\caption{Comparison of SLCAN's probabilistic responses with groundtruth 
annotations on a streaming video. Overall, SLCAN's response (blue curve) exactly 
matches human annotations. The apparent mismatches right before (d) and right after 
(f) mostly correspond to the sections where human annotation workers do not agree unanimously. 
As can be seen in the magnified image of (e), our model can even detect the 
tail of a car on the left lane, which is barely visible, and correctly classify it as {\tt BLOCKED}. 
The graph rarely shows fluctuation without the use of temporal information.
	}\label{fig:prob_blocked}
	\end{figure}

	To verify whether our model performs reliably over time, we examined the probability 
of $P({\tt BLOCKED}|I_L(t))$ on a streaming video containing both free and blocked 
situations (see Fig.~\ref{fig:prob_blocked}). In the video, the left lane is free at 
first and then blocked by stationary obstacles, \eg trees and curb (orange colored 
areas on the graph) and freed again, and finally blocked by passing a car (red colored areas). 
The graph shows that, in most cases, SLCAN's response (blue curve) exactly matches 
human annotations. The apparent mismatches, right before (d) and right after (f), 
mostly correspond to the sections annotated as {\tt UNDEFINED} label, where human 
annotation workers do not agree unanimously. Moreover, it rarely shows fluctuation 
without the use of temporal information.
	
	Also, we tested the Experiment 3 model on streaming video including \texttt{UNDEFINED} labeled images, 
where the images are taken at lane-change motion, roads without lane markings, 
unusual road markings such as crosswalk or intersection. We found that our model 
mostly makes correct prediction regarding the occupancy status of adjacent space 
for such images. We provide video results\footnote{\url{https://github.com/jsgyun/SLCAN}}.
    
	The trained networks can process $\sim$94 frames per second on a PC equipped with an NVIDIA GTX 1080 GPU, making it suitable for real-time 
processing on a consumer-level PC, and even for running on an embedded device at a decent speed.
	
\section{Conclusions and Future Work}
\label{sec:conclusions}
	A novel image based end-to-end decision system for safe lane-change has been 
presented. We showed that even without intermediate steps such as object detection 
and tracking, direct decision from image pixels achieves an acceptable performance 
in terms of accuracy (96.98 \%) and speed (94 fps). We also showed that, without
performance degradation, we can handle left rear view images and flipped right rear 
view images with a single DCNN instead of handling them with separate DCNNs. 
The power of generalization is demonstrated by additional experiment where the DCNN 
trained with images of one type of road (highway) turns out to be able to correctly 
classify images of another type of road (urban roadway). Also, we visualized where 
SLCAN focuses on in an image to make a decision, which shows that SLCAN behaves
like human drivers.

	Our future work will be directed towards better generalization and reliability
of current method. For generalization, we need to expand the current small and 
restricted dataset to a larger one that covers various road conditions such as 
night time scenes, road types, and adverse weather condition. To get more reliable 
lane-change decision, we would exploit temporal information across consecutive 
frames since an approaching vehicle at a distance may or may not be a threat 
depending on the relative speed of that vehicle. To handle this, we plan to design 
an advanced RNN-based SLCAN that feeds on several consecutive image inputs and returns 
more reliable decision.

\section*{Acknowledgement}
\label{sec:ack}
	The authors thank Jongyoon Peck, Jiyoung Jung, Jinhan Lee, Kisung Kim, 
Namil Kim, Sunwook Choi, and Sungjun Choi from NAVER LABS Corp. 
for fruitful discussions on SLCAN dataset collection and annotation works.

\bibliographystyle{ieeetr}
\bibliography{IV2017.bbl}
\end{document}